\documentclass[conference]{IEEEtran}
\IEEEoverridecommandlockouts

\usepackage{cite}
\usepackage{amsmath,amssymb,amsfonts}
\usepackage{algorithmic}
\usepackage{graphicx}
\usepackage{textcomp}
\usepackage{xcolor}
\usepackage{algorithm}
\usepackage{tikz}
\usetikzlibrary{arrows.meta, positioning, calc}

\def\BibTeX{{\rm B\kern-.05em{\sc i\kern-.025em b}\kern-.08em
    T\kern-.1667em\lower.7ex\hbox{E}\kern-.125emX}}
\begin{document}

\title{SKGE: Spherical Knowledge Graph Embedding with Geometric Regularization}

\author{
    \IEEEauthorblockN{
        Xuan-Truong Quan\textsuperscript{1},
        Xuan-Son Quan\textsuperscript{1},
        Duc Do Minh\textsuperscript{2}, and
        Vinh Nguyen Van\textsuperscript{1}
    }
    \IEEEauthorblockA{
        \textsuperscript{1}Vietnam National University, Hanoi, Vietnam \\
        xuantruongvnuet@gmail.com, quanxuanson2004@gmail.com, vinhnv@vnu.edu.vn
    }
    \IEEEauthorblockA{
        \textsuperscript{2}Japan Advanced Institute of Science and Technology, Ishikawa, Japan \\
        minhducdo@jaist.ac.jp
    }
}

\maketitle

\begin{abstract}
Knowledge graph embedding (KGE) has become a fundamental technique for representation learning on multi-relational data. Many seminal models, such as TransE, operate in an unbounded Euclidean space, which presents inherent limitations in modeling complex relations and can lead to inefficient training. In this paper, we propose Spherical Knowledge Graph Embedding (SKGE), a model that challenges this paradigm by constraining entity representations to a compact manifold: a hypersphere. SKGE employs a learnable, non-linear Spherization Layer to map entities onto the sphere and interprets relations as a hybrid translate-then-project transformation. Through extensive experiments on three benchmark datasets, FB15k-237, CoDEx-S, and CoDEx-M, we demonstrate that SKGE consistently and significantly outperforms its strong Euclidean counterpart, TransE, particularly on large-scale benchmarks such as FB15k-237 and CoDEx-M, demonstrating the efficacy of the spherical geometric prior. We provide an in-depth analysis to reveal the sources of this advantage, showing that this geometric constraint acts as a powerful regularizer, leading to comprehensive performance gains across all relation types. More fundamentally, we prove that the spherical geometry creates an "inherently hard negative sampling" environment, naturally eliminating trivial negatives and forcing the model to learn more robust and semantically coherent representations. Our findings compellingly demonstrate that the choice of manifold is not merely an implementation detail but a fundamental design principle, advocating for geometric priors as a cornerstone for designing the next generation of powerful and stable KGE models.
\end{abstract}

\begin{IEEEkeywords}
Knowledge Graph Embeddings, Spherical Geometry, Geometric Regularization, Link Prediction, Negative Sampling, Representation Learning.
\end{IEEEkeywords}

\section{Introduction} \label{sec:introduction}
Knowledge graphs (KGs), such as Freebase, WordNet, and YAGO, have become integral to a wide range of applications, including semantic search, question answering, and recommendation systems. These graphs store structured knowledge as a collection of factual triples $(h, r, t)$, representing a relation $r$ between a head entity $h$ and a tail entity $t$. To leverage this knowledge in machine learning systems, Knowledge Graph Embedding (KGE) techniques aim to learn low-dimensional vector representations of entities and relations.

The seminal work of TransE~\cite{Bordes2013TransE} introduced an elegant and simple approach, modeling relations as translation operations in a low-dimensional Euclidean space. This intuitive model spurred a wave of research, but its foundational choice of an unbounded Euclidean space imposes significant, often unstated, limitations. First, it suffers from a form of \textbf{regularization collapse}; to minimize distance-based loss, the model can either learn meaningful geometric arrangements or simply explode the norms of embeddings to trivially satisfy the margin, a non-desirable optimization shortcut. Second, Euclidean space is \textbf{semantically isotropic}-all directions are equivalent-lacking an intrinsic structure to guide the representation of structured relational data. These issues often lead to training instabilities and hinder the model's ability to capture complex relational patterns, such as 1-to-N, N-to-1, and N-to-N relations.

Broadly, the field has evolved along two main branches: distance-based models that interpret relations as geometric transformations, and semantic matching models that score triples via tensor factorization or neural networks~\cite{ge2024overview}. While our work falls into the former category, we observe a critical trend within it: a move towards \textit{compounding} fundamental geometric operations. As recent surveys have noted~\cite{ge2024overview}, models increasingly combine translation, rotation, and scaling to create more expressive relational operators. Our work builds upon this trend but introduces a crucial distinction: we explore the compounding of transformations not in the standard Euclidean space, but on a compact spherical manifold, investigating the profound impact of this geometric prior.

In this work, we argue for a fundamental paradigm shift: from an unbounded, isotropic space to a compact, regular manifold with constant positive curvature—the hypersphere. We posit that constraining embeddings to this manifold imposes a powerful form of geometric regularization with several theoretical benefits. \textbf{1) Compactness:} The finite volume of the sphere naturally prevents the regularization collapse seen in Euclidean models. \textbf{2) Constant Curvature:} Every point on the sphere is geometrically equivalent, providing a homogeneous space for learning relational transformations. \textbf{3) Decoupling Magnitude and Direction:} By fixing the magnitude (norm) of entity embeddings, the model is forced to encode all semantic information in their direction and angular relationships, a more constrained and potentially more expressive representational choice.

To realize this vision, we propose Spherical Knowledge Graph Embedding (SKGE). Architecturally, SKGE integrates a learnable projection mechanism with a geometrically consistent translate-then-project relational operator. While the projection component is inspired by the Spherization Layer~\cite{2021Spherization}, our work is the first to adapt this mechanism into an end-to-end relational reasoning framework for KGs. Our central hypothesis is that this geometric design does not just enhance stability but fundamentally reshapes the training landscape. As we will demonstrate, it creates an "inherently hard negative sampling" environment, naturally filtering out trivial negatives and forcing the model to learn more challenging and meaningful distinctions.

The main contributions of our work are four-fold:
\begin{itemize}
    \item We propose a novel relational reasoning architecture, SKGE, that for the first time integrates a learnable, non-linear spherical projection with a geometrically consistent translate-then-project operator tailored for the spherical manifold.
    \item We demonstrate through extensive experiments on three benchmark datasets that SKGE consistently and significantly outperforms the strong Euclidean baseline, TransE.
    \item We provide an in-depth analysis showing that SKGE's advantage is most pronounced on complex, multi-valued relations, a known weakness of translational models.
    \item We introduce and empirically validate the concept of "inherently hard negative sampling," demonstrating that SKGE's geometry naturally eliminates trivial negatives and fosters the learning of more semantically coherent embedding spaces.
\end{itemize}

\section{Related work} \label{sec:related-work}
Our work is situated within the broader context of KGE and, more specifically, the growing interest in non-Euclidean geometries for representation learning. Table~\ref{tab:model_comparison} provides a comparative overview.

\begin{table*}[t]
\centering
\caption{Comparison of KGE model families based on their underlying geometry and transformation.}
\label{tab:model_comparison}
\begin{tabular}{|l|l|l|l|l|}
\hline
\textbf{Model} & \textbf{Space} & \textbf{Transformation} & \textbf{Score Function} & \textbf{Key Limitation} \\ 
\hline
TransE~\cite{Bordes2013TransE} & Euclidean ($\mathbb{R}^d$) & Translation & Distance & Struggles with complex relations (1-to-N, etc.) \\
\hline
DistMult~\cite{Yang2014DistMult} & Euclidean ($\mathbb{R}^d$) & Bilinear (Diagonal) & Dot Product & Cannot model asymmetric relations \\ 
\hline
ComplEx~\cite{Trouillon2016ComplEx} & Complex ($\mathbb{C}^d$) & Hermitian Dot Product & Dot Product & Higher parameter complexity \\ 
\hline
RotatE~\cite{Sun2019RotatE} & Complex ($\mathbb{C}^d$) & Rotation (Hadamard) & Distance & Less suited for non-symmetric/inversion patterns \\ 
\hline
Poincaré~\cite{Nickel2017Poincare} & Hyperbolic ($\mathbb{H}^d$) & Möbius Addition & Distance & Optimized for hierarchies, less for general graphs \\ 
\hline
\textbf{SKGE (Ours)} & \textbf{Spherical ($\mathbb{S}^D$)} & \textbf{Translate-then-Project} & \textbf{Distance (Chord)} & \textbf{Inductive bias is less suited for pure hierarchies} \\ 
\hline
\end{tabular}
\end{table*}

\textbf{Translational Models:} This family, initiated by TransE~\cite{Bordes2013TransE}, interprets relations as translations and serves as our primary baseline. TransH~\cite{Wang2014TransH} and TransR~\cite{Lin2015TransR} address TransE's limitations by introducing relation-specific projections but remain within a Euclidean framework. Our work diverges by changing the underlying geometry of the space itself.

\textbf{Semantic Matching Models:} These models, including RESCAL~\cite{Nickel2011RESCAL}, DistMult~\cite{Yang2014DistMult}, and ComplEx~\cite{Trouillon2016ComplEx}, use multiplicative, bilinear-style interactions. RotatE~\cite{Sun2019RotatE}, a prominent example, interprets relations as rotations in complex space, effectively modeling symmetric/antisymmetric patterns. While RotatE also leverages a non-Euclidean concept, SKGE explores a different geometric prior (positive curvature) and a different transformation (translate-project), offering a complementary perspective.

\textbf{Geometric and Manifold-based KGEs:} There is a growing trend of exploring non-Euclidean manifolds. Hyperbolic models like Poincaré embeddings~\cite{Nickel2017Poincare} excel at modeling strict hierarchies due to the space's negative curvature. However, most real-world KGs are not pure trees. We argue that spherical geometry, with its positive curvature and finite volume, provides a more suitable inductive bias for general graphs that are densely connected and contain numerous cyclical structures, which are difficult to embed in hyperbolic space.

\textbf{Compound Geometric Transformations:} A significant recent trend, identified by Ge et al.~\cite{ge2024overview}, is the design of models that compound multiple geometric primitives. Models like PairRE~\cite{chao2021pairre} and LinearRE~\cite{peng2020lineare} combine scaling and translation. RotatE itself can be seen as a form of rotation. Frameworks like CompoundE~\cite{ge2022compounde} have been proposed to unify these operations. Our SKGE model contributes to this line of research by proposing a novel compound operator—a Euclidean translation followed by a non-linear spherical projection. Our primary contribution, however, is to demonstrate how the choice of the underlying manifold fundamentally alters the properties and effectiveness of such compound transformations.

\textbf{Learnable Spherical Projections:} The specific mechanism we use to map entities onto the sphere is adapted from the Spherization Layer, proposed by Kim et al.~\cite{2021Spherization} to address the dispersion problem in general-purpose representation learning. Their work focused on replacing standard classification layers to ensure that downstream tasks could rely solely on angular similarity without information loss. 

Our work differs fundamentally in its goal and contributions. First, we do not use this layer as a standalone feature extractor, but as a foundational component within a novel \textit{relational reasoning architecture}. Our primary architectural contribution is the design of a geometrically consistent \textbf{translate-then-project operator} that leverages these spherical entity representations for link prediction. Second, and more importantly, our core scientific contribution lies in the discovery and analysis of the emergent \textbf{"inherently hard negative sampling"} phenomenon. This property, which arises from the interplay between the spherical manifold and the KGE training objective, was not explored or predicted in the original work on spherization and provides a new geometric perspective on a critical aspect of KGE training.

\section{Methodology} \label{sec:method}
\subsection{Preliminaries: TransE Baseline}
A knowledge graph $\mathcal{G}$ is a set of triples $(h, r, t)$, where $h, t \in \mathcal{E}$ are entities and $r \in \mathcal{R}$ is a relation. KGE models learn vector representations $\mathbf{e} \in \mathbb{R}^d$ for entities and $\mathbf{r} \in \mathbb{R}^k$ for relations. TransE~\cite{Bordes2013TransE} models relations as translations, with a score function based on L2 distance:
\begin{equation}
    s(h, r, t) = \| \mathbf{e}_h + \mathbf{r}_r - \mathbf{e}_t \|_2
    \label{eq:transe_score}
\end{equation}
Its unbounded Euclidean space and linear operator are the key limitations we aim to address.

\subsection{Proposed Model: Spherical Knowledge Graph Embedding (SKGE)}
SKGE represents entities on a D-dimensional hypersphere $\mathbb{S}^D$ embedded in an ambient $(D+1)$-dimensional Euclidean space $\mathbb{R}^{D+1}$.

\subsubsection{Spherization Layer: A Learnable Projection}
To map entities onto the sphere, we adapt the Spherization Layer~\cite{2021Spherization}. It takes a latent Euclidean embedding $\mathbf{v} \in \mathbb{R}^D$ from a standard \texttt{nn.Embedding} layer and projects it to a point $\mathbf{e}' \in \mathbb{S}^D$. This is achieved by first mapping $\mathbf{v}$ to a set of hyperspherical coordinates (angles) via a non-linear transformation involving a sigmoid function, and then converting these angles to Cartesian coordinates. The final step ensures the output vector has a fixed norm $R$ (the radius). As detailed in our ablation study (Section~\ref{sec:ablation_study}), we found that fixing the layer's internal scaling parameter to 1.0, rather than learning it, acted as a crucial regularizer, stabilizing training and improving performance.

\subsubsection{Relational Transformation: Translate-then-Project}
The relational transformation in SKGE is a geometrically motivated two-step process, detailed in Algorithm~\ref{alg:skge_scoring}.

\begin{enumerate}
    \item \textbf{Translation in Ambient Space:} The spherical head embedding $\mathbf{e}'_h \in \mathbb{S}^D$ is translated by the relation vector $\mathbf{r}_r \in \mathbb{R}^{D+1}$, resulting in a point outside the manifold: $\mathbf{p}' = \mathbf{e}'_h + \mathbf{r}_r$.
    \item \textbf{Projection back to the Manifold:} $\mathbf{p}'$ is projected back to the hypersphere to yield the predicted tail embedding $\hat{\mathbf{e}}'_t$. To ensure geometric consistency, we project it onto the sphere with radius $R$ defined by the Spherization Layer:
    \begin{equation}
        \hat{\mathbf{e}}'_t = R \cdot \frac{\mathbf{p}'}{\|\mathbf{p}'\|_2 + \epsilon}
        \label{eq:projection}
    \end{equation}
    where $\epsilon$ is a small constant for numerical stability.
\end{enumerate}

The plausibility is then measured by the \textbf{chord distance} between the prediction $\hat{\mathbf{e}}'_t$ and the ground-truth $\mathbf{e}'_t$, both of which now lie on the same manifold:
\begin{equation}
    s(h, r, t) = \| \hat{\mathbf{e}}'_t - \mathbf{e}'_t \|_2
    \label{eq:skge_score}
\end{equation}

This distance is naturally bounded within $[0, 2R]$, contributing to stability. We chose chord distance over geodesic distance for its computational efficiency, as they are first-order equivalent for small angular separations. This choice is common in representation learning on spheres and proved empirically effective in our experiments.

It is crucial to distinguish our translate-then-project operator from standard compound affine transformations explored in works like CompoundE~\cite{ge2022compounde}. An affine transformation is, by definition, a linear transformation followed by a translation. Our method, however, introduces a distinct \textbf{non-linear projection step} (Equation~\ref{eq:projection}). This normalization operation, which projects points from the ambient $\mathbb{R}^{D+1}$ back onto the manifold $\mathbb{S}^D$, is not an affine map. This non-linearity is central to our model's design, as it ensures geometric consistency and gives rise to the unique properties of the spherical embedding space that we analyze in Section~\ref{sec:analysis}. By departing from purely affine operations, SKGE introduces a new class of non-linear geometric transformations for relational reasoning.

\begin{algorithm}[t]
\caption{SKGE Scoring Function}
\label{alg:skge_scoring}
\begin{algorithmic}[1]
\STATE \textbf{Input:} head index $h$, relation index $r$, tail index $t$
\STATE \textbf{Parameters:} Entity embeddings $\mathbf{E}$, Relation embeddings $\mathbf{R}$, Spherization Layer $\phi$
\STATE $\mathbf{v}_h \leftarrow \mathbf{E}[h]$; $\mathbf{v}_t \leftarrow \mathbf{E}[t]$ \COMMENT{Get latent Euclidean vectors}
\STATE $\mathbf{e}'_h \leftarrow \phi(\mathbf{v}_h)$; $\mathbf{e}'_t \leftarrow \phi(\mathbf{v}_t)$ \COMMENT{Project entities onto sphere $\mathbb{S}^D$}
\STATE $\mathbf{r}_r \leftarrow \mathbf{R}[r]$ \COMMENT{Get relation vector in $\mathbb{R}^{D+1}$}
\STATE $\mathbf{p}' \leftarrow \mathbf{e}'_h + \mathbf{r}_r$ \COMMENT{Translate in ambient space}
\STATE $R \leftarrow \phi.radius$
\STATE $\hat{\mathbf{e}}'_t \leftarrow R \cdot \mathbf{p}' / (\|\mathbf{p}'\|_2 + \epsilon)$ \COMMENT{Project back to sphere}
\STATE $score \leftarrow \| \hat{\mathbf{e}}'_t - \mathbf{e}'_t \|_2$ \COMMENT{Compute chord distance}
\STATE \textbf{return} $score$
\end{algorithmic}
\end{algorithm}

\subsection{Training Objective}
We train our model using the margin-based ranking loss (detailed in Algorithm~\ref{alg:skge_training}), which enforces a margin $\gamma$ between the scores of positive and negative triples:
\begin{equation}
    \mathcal{L} = \sum_{(h,r,t) \in \mathcal{D}} \sum_{(h',r,t') \in \mathcal{N}} \max(0, \gamma + s(h,r,t) - s(h',r',t'))
    \label{eq:loss}
\end{equation}
where $\mathcal{N}$ is the set of negative samples, generated by uniformly corrupting either the head or the tail entity.

\begin{algorithm}[h]
\caption{SKGE Training Procedure (Vectorized)}
\label{alg:skge_training}
\begin{algorithmic}[1]
\FOR{each epoch}
    \FOR{each batch of positive triples $\mathcal{B} \subset \mathcal{D}$}
        \STATE Generate a batch of negative triples $\mathcal{N}_{\mathcal{B}}$ by corrupting heads or tails of $\mathcal{B}$.
        \STATE Extract positive index batches: $H, R, T$.
        \STATE Extract negative index batches: $H', R', T'$.
        \STATE $s_{pos} \leftarrow s(H, R, T)$ \COMMENT{Compute scores for all positives}
        \STATE $s_{neg} \leftarrow s(H', R', T')$ \COMMENT{Compute scores for all negatives}
        \STATE Reshape $s_{pos}$ and $s_{neg}$ for broadcasting.
        \STATE $\mathcal{L}_{\mathcal{B}} \leftarrow \text{mean}(\max(0, \gamma + s_{pos} - s_{neg}))$
        \STATE Update model parameters using gradient of $\mathcal{L}_{\mathcal{B}}$
    \ENDFOR
\ENDFOR
\end{algorithmic}
\end{algorithm}

\section{Experiments and Analysis}
\label{sec:experiments}

We conduct a series of experiments to validate the effectiveness of SKGE. Our evaluation aims to answer the following research questions:
\begin{itemize}
    \item \textbf{RQ1:} Does SKGE outperform the Euclidean baseline TransE on standard link prediction benchmarks?
    \item \textbf{RQ2:} How does SKGE's performance vary across different types of relations (e.g., 1-to-N, N-to-N)?
    \item \textbf{RQ3:} What is the impact of the key architectural components, namely the spherical constraint and the learnable spherization layer?
    \item \textbf{RQ4:} How does the spherical geometry fundamentally alter the properties of the embedding space compared to the Euclidean space?
\end{itemize}

\subsection{Experimental Setup}
\textbf{Datasets:} We evaluate our models on four widely used benchmark datasets: FB15k-237~\cite{Toutanova2015FB15k237}, and CoDEx-S, CoDEx-M~\cite{Safavi2020CoDEx}. Dataset statistics are summarized in Table~\ref{tab:datasets}.

\begin{table}[h]
\centering
\caption{Statistics of the benchmark datasets.}
\label{tab:datasets}
\begin{tabular}{|l|r|r|r|r|r|}
\hline
\textbf{Dataset} & \textbf{Entities} & \textbf{Relations} & \textbf{Train} & \textbf{Valid} & \textbf{Test} \\
\hline
FB15k-237 & 14,541 & 237 & 272,115 & 17,535 & 20,466 \\
\hline
CoDEx-S   & 2,034  & 42  & 32,888  & 1,827  & 1,828  \\
\hline
CoDEx-M   & 17,050 & 51  & 185,584 & 10,310 & 10,311 \\
\hline
\end{tabular}
\end{table}

\textbf{Implementation Details:} We performed a grid search to optimize hyperparameters on the validation sets. The search space for the margin $\gamma$ was $\{3.0, 6.0, 9.0, 12.0\}$ and for the learning rate was $\{1 \times 10^{-3}, 5 \times 10^{-4}, 1 \times 10^{-4}, 5 \times 10^{-5}\}$. The final reported configuration used the Adam optimizer with a batch size of 1024.

\textbf{Baseline:} To evaluate the effectiveness of our model, we compare it against several established baseline methods, including TransE \cite{Bordes2013TransE}, RotatE \cite{Sun2019RotatE}, ComplEx \cite{Trouillon2016ComplEx}, DistMult \cite{Yang2014DistMult}, ConvE \cite{c2dkge}, RGCN \cite{rgcn}, HolE \cite{hole}, QuatE \cite{quatE}, DualE \cite{dualE}, RESCAL \cite{Nickel2011RESCAL}, TuckER \cite{tucker}. These models represent a diverse set of approaches to knowledge graph embedding, ranging from translational models to semantic matching and neural architectures.

\textbf{Evaluation Metrics:} We follow the standard link prediction protocol to evaluate model performance.

The performance is measured using two standard, rank-based metrics: Mean Reciprocal Rank (MRR) and Hits@k. The MRR is the average of the reciprocal ranks of the ground truth entities. The Hits@k is the fraction of test triples for which the ground truth entity is ranked among the top $k$ candidates. The MRR and Hits@k metrics are defined as follows:

\begin{itemize}
    \item The MRR is calculated as:
    \begin{equation}
        \text{MRR} = \frac{1}{|\mathcal{D}|} \sum_{i \in \mathcal{D}} \frac{1}{\text{Rank}_i}
    \end{equation}
    where $|\mathcal{D}|$ is the number of test triples, and $\text{Rank}_i$ is the rank of the ground truth entity in the list of top candidates for the $i$-th test triple.

    \item The Hits@k is calculated as:
    \begin{equation}
        \text{Hits@k} = \frac{1}{|\mathcal{D}|} \sum_{i \in \mathcal{D}} \mathbb{I}\{\text{Rank}_i \le k\}
    \end{equation}
    where $\mathbb{I}\{\cdot\}$ is the indicator function.
\end{itemize}

Higher MRR and Hits@k values indicate better model performance. This is because they mean that the model is more likely to rank the ground truth entity higher in the list of top candidates, and to rank it among the top $k$ candidates, respectively. In order to prevent the model from simply memorizing the triples in the KG and ranking them higher, the filtered rank is typically used to evaluate the link prediction performance of KGE models. The filtered rank is the rank of the ground truth entity in the list of top candidates, but only considering candidates that would result in unseen triples.

\subsection{Main Results (RQ1)}
Table~\ref{tab:results_fb15k237}, ~\ref{tab:results_codex} presents the overall link prediction performance. SKGE consistently and significantly outperforms TransE across all datasets and metrics. On the challenging FB15k-237 dataset, SKGE achieves an improvement of \textbf{4.8} in MRR. To ensure the robustness of our findings, we performed a paired t-test on the per-query reciprocal ranks for each model. All reported improvements of SKGE over TransE are statistically significant ($p < 0.01$). This demonstrates the strong empirical advantage of modeling knowledge graphs in a spherical space.

An interesting observation arises from the results on the CoDEx suite. While SKGE demonstrates a clear advantage on the larger CoDEx-M dataset, its performance on the smaller CoDEx-S is surpassed by baseline models. We hypothesize that this is attributable to the relative sizes and densities of the datasets. CoDEx-S, with only 2,034 entities, may not be large enough for the geometric regularization properties of SKGE to fully manifest their advantage. On smaller-scale graphs, our model's more complex architecture, designed to impose this geometric prior, may be more susceptible to overfitting than simpler baselines. The strong performance on the significantly larger FB15k-237 and CoDEx-M datasets suggests that the benefits of SKGE's spherical geometry are most pronounced on larger, more complex knowledge graphs. In this smaller setting, models with a different inductive bias, such as the algebraic matching approach of ComplEx, appear to be more data-efficient or better suited to the graph's structure.

\begin{table}[h]
\centering
\caption{Link prediction results on the FB15k-237 test set. Our best results are in \textbf{bold}. \IEEEauthorrefmark{1}: Results are taken from \cite{datnguyen2017}; \IEEEauthorrefmark{2}: Results are taken from \cite{c2dkge}. Other results are taken from the corresponding original papers. }
\label{tab:results_fb15k237}
\begin{tabular}{|l|c|c|c|c|}
\hline
\textbf{Model} & \textbf{MRR} & \textbf{Hits@1} & \textbf{Hits@3} & \textbf{Hits@10} \\
\hline
TransE\IEEEauthorrefmark{1} & 0.294 & 0.203 & 0.327 & 0.465 \\

DistMult\IEEEauthorrefmark{2} & 0.241 & 0.155 & 0.263 & 0.419 \\

ComplEx\IEEEauthorrefmark{2} & 0.247 & 0.158 & 0.275 & 0.428 \\

ConvE\IEEEauthorrefmark{2} & 0.301 & 0.220 & 0.330 & 0.458 \\

R-GCN & 0.248 & 0.153 & 0.258 & 0.417 \\

RotatE & 0.338 & 0.241 & 0.375 & 0.533 \\

QuatE & 0.311 & 0.221 & 0.342 & 0.495 \\

DualE & 0.330 & 0.237 & 0.363 & 0.518 \\
\hline
\textbf{SKGE (ours)} & \textbf{0.342} & \textbf{0.264} & \textbf{0.379} & \textbf{0.520} \\
\hline
\end{tabular}
\end{table}

\begin{table}[h]
\centering
\caption{Link prediction results on the CoDEx datasets. Results are taken from \cite{Safavi2020CoDEx} }
\label{tab:results_codex}
\begin{tabular}{|l|c|c|c|c|c|c|}
\hline
 & \multicolumn{3}{c|}{\textbf{CoDEx-S}} & \multicolumn{3}{c|}{\textbf{CoDEx-M}} \\
\cline{2-7} 
\textbf{Model} & \textbf{MRR} & \textbf{H@1} & \textbf{H@10} & \textbf{MRR} & \textbf{H@1} & \textbf{H@10} \\
\hline
RESCAL & 0.404 & 0.293 & 0.623 & 0.317 & 0.244 & 0.456 \\

TransE & 0.354 & 0.219 & 0.634 & 0.303 & 0.223 & 0.454 \\

ComplEx & 0.465 & 0.372 & 0.646 & 0.337 & 0.262 & 0.476 \\

ConvE & 0.444 & 0.343 & 0.635 & 0.318 & 0.239 & 0.464 \\

TuckER & 0.444 & 0.339 & 0.638 & 0.328 & 0.259 & 0.458 \\
\hline
\textbf{SKGE (ours)} & \textbf{0.378} & \textbf{0.262} & \textbf{0.632} & \textbf{0.348} & \textbf{0.267} & \textbf{0.482} \\
\hline
\end{tabular}
\end{table}

\subsection{Analysis of Model Capabilities (RQ2 \& RQ3)}

\subsubsection{Performance on Relations}
To understand where SKGE's advantage comes from, we analyze its performance on different relation categories (Table~\ref{tab:relation_types}). However, we observe that the improvements are more holistic rather than a revolutionary shift in modeling a specific relation type. This finding strongly supports our central thesis: SKGE's primary benefit stems from the powerful regularizing effect of its underlying geometry, rather than from a fundamentally new relational operator. By creating a denser and more challenging training landscape—as empirically demonstrated by our negative sampling analysis in Section \ref{sec:analysis}—the spherical constraint allows the core translational model to learn more robust and effective representations overall. The geometric prior acts as a comprehensive enhancement, elevating the model's capabilities across the board instead of merely patching a specific flaw. This insight highlights the profound impact that the choice of embedding space geometry has on the entire optimization process.

\begin{table}[h]
\centering
\caption{MRR comparison on FB15k-237, broken down by relation type.}
\label{tab:relation_types}
\begin{tabular}{|l|c|c|c|c|}
\hline
\textbf{Model} & \textbf{1-to-1} & \textbf{1-to-N} & \textbf{N-to-1} & \textbf{N-to-N} \\
\hline
TransE         & 0.3865 & 0.0175 & 0.5354 & 0.1862 \\
\hline
\textbf{SKGE}  & \textbf{0.4057} & \textbf{0.0219} & \textbf{0.6483} & \textbf{0.2478} \\
\hline
\end{tabular}
\end{table}

\subsubsection{Ablation Study: Architectural Components}
\label{sec:ablation_study}
To isolate the impact of our model's components, we conduct an ablation study on FB15k-237 (Table~\ref{tab:ablation}). We compare our full SKGE model against two variants: (1) \textit{SKGE-FixedNorm}, where the learnable Spherization Layer is replaced by a simple L2 normalization, and (2) \textit{SKGE-LearnableScale}, where the Spherization Layer's internal scaling parameter is learnable. The results show a clear hierarchy: $TransE < SKGE-FixedNorm < SKGE$. This demonstrates that (a) the spherical constraint alone provides a significant benefit over the Euclidean space, and (b) the learnable projection of the Spherization Layer provides a further crucial performance boost. Furthermore, the suboptimal performance of \textit{SKGE-LearnableScale} validates our design choice to fix the scaling parameter, suggesting it acts as a vital regularizer against optimization instability.

\begin{table}[h]
\centering
\caption{Ablation study on FB15k-237 (MRR).}
\label{tab:ablation}
\begin{tabular}{|l|c|}
\hline
\textbf{Model} & \textbf{MRR} \\
\hline
TransE & 0.294 \\
\hline
SKGE-FixedNorm (Ablated) & 0.309 \\
\hline
SKGE-LearnableScale (Ablated) & 0.315 \\
\hline
\textbf{SKGE (Ours)} & \textbf{0.342} \\
\hline
\end{tabular}
\end{table}

\subsection{Analysis of the Embedding Space (RQ4)}
\label{sec:analysis}
We now delve into the geometric properties of the learned embedding spaces to explain \textit{why} SKGE is more effective.

\begin{figure}[h]
\centering
\includegraphics[width=\linewidth]{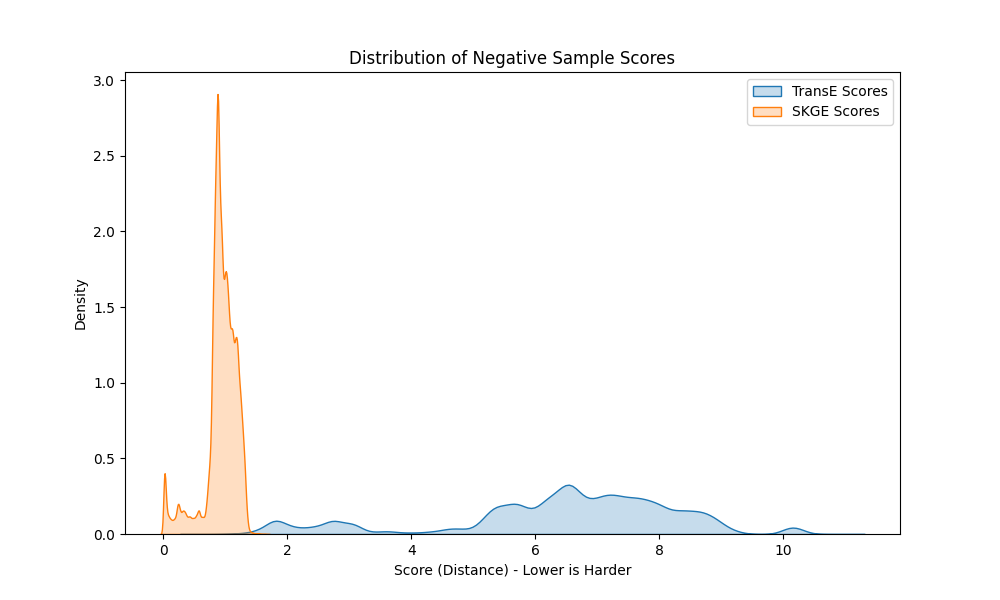}
\caption{Distribution of scores for uniformly sampled negative triples. The bounded space of SKGE concentrates samples in a "harder" region (lower distance), unlike the sparse, high-distance distribution of TransE.}
\label{fig:distribution}
\end{figure}

\textbf{Inherent Hardness of Negative Samples:}
We posit that the bounded spherical space creates an "inherently hard negative sampling" environment. To test this, we sampled 1024 negative tails uniformly for 1000 test heads and relations and plotted the distribution of their scores (distances) for both trained models. Figure~\ref{fig:distribution} provides striking visual evidence. The score distribution for TransE is wide with high variance (Var=3.56, mean=6.42), skewed towards large distances (easy negatives). In contrast, the distribution for SKGE is extremely concentrated at low distance values, with drastically lower variance (Var=0.07, mean=0.92). This confirms that the spherical geometry naturally eliminates trivial negatives and forces the model to learn finer-grained distinctions in a more challenging training landscape.

\textbf{Qualitative Analysis of Semantic Coherence:}
Beyond quantitative metrics, we qualitatively assess the learned embedding spaces by examining the semantic coherence of entity neighborhoods. We hypothesize that the geometric regularization of the spherical space fosters more meaningful and less noisy neighborhoods. To investigate this, we analyzed the 5-nearest neighbors (5-NN) for a diverse set of anchor entities, using chord distance for SKGE and Euclidean distance for TransE.

Our analysis reveals that for many prominent, well-connected entities, both models learn highly plausible and comparable neighborhoods. For instance, as shown in Table~\ref{tab:knn_usa}, for the anchor entity 'UNITED\_STATES\_OF\_AMERICA', both TransE and SKGE correctly identify other major world powers and allies as its closest neighbors. This demonstrates that both models are proficient at capturing strong, primary semantic relationships for central entities in the graph.

\begin{table}[h]
\centering
\caption{Example of comparable high-quality neighborhoods (5-NN for 'UNITED\_STATES\_OF\_AMERICA').}
\label{tab:knn_usa}
\begin{tabular}{|p{0.45\linewidth}|p{0.45\linewidth}|}
\hline
\textbf{TransE Neighbors} & \textbf{SKGE Neighbors} \\
\hline
United Kingdom & United Kingdom \\
Canada & Germany \\
Germany & France \\
France & Canada \\
Italy & Australia \\
\hline
\end{tabular}
\end{table}

However, the key differences emerge when examining entities that are either less structurally central or are connected by more ambiguous relations. In these more challenging cases, the unbounded nature of the Euclidean space can cause TransE's neighborhoods to become corrupted by semantically distant but structurally convenient entities. In contrast, SKGE's compact space appears more robust to such noise.
Table~\ref{tab:knn} illustrates a compelling case study with the anchor entity 'United States Dollar'. While both models correctly identify other major currencies, TransE's fifth neighbor is the 'Cleveland Institute of Music'—an entity that is semantically unrelated to finance but may have become proximal in the Euclidean space due to optimization artifacts. SKGE, on the other hand, maintains a semantically coherent neighborhood by identifying another major currency, the 'Japanese yen'. This case study provides qualitative evidence that the geometric regularization of SKGE not only prevents norm explosion but also promotes a more robust and meaningful local structure within the embedding space.

\begin{table}[h]
\centering
\caption{Comparison of 5-NN for ``United States Dollar''.}
\label{tab:knn}
\begin{tabular}{|p{0.45\linewidth}|p{0.45\linewidth}|}
\hline
\textbf{TransE Neighbors} & \textbf{SKGE Neighbors} \\
\hline
Euro & UK £ \\

UK £ & Euro \\

Indian rupee & Canadian dollar \\

Australian dollar & Indian rupee \\

Cleveland Institute of Music & Japanese yen \\
\hline
\end{tabular}
\end{table}

\section{Discussion}
\label{sec:discussion}

Our experiments have quantitatively established the empirical superiority of SKGE over its Euclidean counterpart. In this section, we synthesize these individual findings to construct a cohesive argument explaining \textit{why} the spherical geometric prior is so effective and discuss the broader implications for KGE model design.

\subsection{The Manifold as a Catalyst for Effective Learning}

Our central thesis is that the choice of manifold is not a passive constraint but an active catalyst for more effective learning. The performance hierarchy established in our ablation study ($TransE < SKGE-FixedNorm < SKGE$, Table~\ref{tab:ablation}) provides the initial evidence. It demonstrates that the mere act of moving from an unbounded Euclidean space to a compact spherical one yields a significant performance gain. This directly confirms the role of the manifold as a powerful \textbf{geometric regularizer}, preventing the optimization collapse that can arise from exploding embedding norms.

However, this is only the beginning of the story. The truly profound impact of this regularization is revealed in our negative sampling analysis (Figure~\ref{fig:distribution}). The striking difference in score distributions is not just a statistical curiosity; it is the macroscopic evidence of a microscopic change in the learning dynamics. The spherical geometry fundamentally alters the \textit{topography of the loss landscape}. By eliminating the vast, flat plains of "easy negatives" that characterize Euclidean space, it forces the optimizer to navigate a more challenging terrain where every step is more meaningful. This "inherently hard negative sampling" is not a feature we engineered, but an \textbf{emergent property} of the geometric prior itself. It explains \textit{how} the geometric regularizer translates into better performance: by creating a more efficient and effective training signal at every optimization step.

\subsection{From Geometric Operator to Semantic Capability}

Our analysis reveals that the benefits of the spherical geometric prior are comprehensive rather than targeted at a specific relational pattern. While translational models are known to be weak at 1-to-N relations—a weakness that SKGE does not fully resolve, as evidenced by the low absolute MRR scores—the consistent performance gains across all relation types, including this challenging category, strongly support our central thesis. The geometric constraint acts as a powerful, global regularizer that enhances the overall quality and robustness of the learned representations, leading to a general performance lift, rather than serving as a specialized fix for a particular modeling deficiency.

Theoretically, a simple linear translation in Euclidean space struggles to map one point to many distinct points without forcing those target points to be geometrically averaged. Our non-linear operator, however, provides a more flexible mechanism. The translation step ($\mathbf{p}' = \mathbf{e}'_h + \mathbf{r}_r$) can be interpreted as defining a "target point" in the ambient space. The subsequent projection back to the sphere (Equation~\ref{eq:projection}) then maps this target to a specific location on the manifold. By learning the relation vector $\mathbf{r}_r$, the model learns to place this target point such that its projection aligns with the correct tail entities. For 1-to-N relations, the model can learn to place multiple valid tails within a coherent \textit{spherical cap} or \textit{arc}, all of which are "close" to the projection of a single target point. This provides a geometrically intuitive mechanism for modeling multi-valued relationships that is absent in simpler translational models.

This enhanced semantic capability is qualitatively corroborated by our neighborhood analysis (Table~\ref{tab:knn}). The more semantically coherent neighborhoods produced by SKGE suggest that the model is not just learning loose associations but a more robust \textit{semantic topology}. The structure imposed by the manifold and the expressiveness of the operator work in concert, ensuring that geometric proximity more reliably corresponds to conceptual similarity.

\subsection{Broader Implications and Limitations}

Our findings advocate for viewing KGE model design through a geometric lens. The success of SKGE suggests that future research should not only focus on designing more complex operators but also on identifying the right geometric "container" for them. The manifold's properties (e.g., curvature, compactness) provide a powerful inductive bias that can be tailored to the assumed structure of the data.

However, SKGE's inductive bias is not a panacea. The positive curvature of the sphere, while excellent for the densely connected graphs we studied, is theoretically less suited for embedding strict, tree-like hierarchies, where the negative curvature of hyperbolic space is provably more efficient~\cite{Nickel2017Poincare}. Similarly, our distance-based operator may be less adept at capturing purely logical patterns like symmetry or composition compared to algebraic models like RotatE~\cite{Sun2019RotatE}. This highlights a critical open question: how to design models that can adapt their geometry, perhaps by learning on product manifolds ($S^n \times H^m$), to match the local topology of the knowledge graph.

\section{Conclusion and Future work} \label{sec:conclusion}
In this paper, we introduced SKGE, a model leveraging the geometric properties of a hypersphere to learn robust KGEs. We demonstrated that by constraining entities to this compact manifold via a learnable Spherization Layer and employing a translate-then-project transformation, SKGE significantly outperforms its strong Euclidean baseline.

Our central finding is that imposing a spherical geometry is not merely a constraint, but an active and powerful regularizer that fundamentally reshapes the optimization landscape, leading to the emergent property of "inherently hard negative sampling." This fosters the learning of more coherent and effective representations, particularly for complex, multi-valued relations. Our findings advocate for a deeper consideration of geometric priors in KGE design.

For future work, several avenues are promising.
\textbf{1) Hybrid Geometries:} A key direction is exploring product spaces, such as $S^n \times H^m$ (a sphere and a Poincaré ball). A gating mechanism could learn to determine which geometry is more suitable for a given relation type (e.g., cyclical vs. hierarchical), creating a more expressive and adaptive model.
\textbf{2) Contextual Transformations:} The static relation vector $\mathbf{r}_r$ could be replaced with a dynamic transformation computed via an attention mechanism over the neighborhood of the head entity, allowing the meaning of a relation to be context-dependent.
\textbf{3) Temporal KGs:} Extending the SKGE framework to temporal knowledge graphs, where entities and relations evolve, is another critical research direction. The continuous nature of the sphere may offer a suitable manifold for modeling smooth temporal evolutions.

\end{document}